\documentclass[runningheads]{llncs}
\usepackage{graphicx}
\usepackage{float}
\usepackage{booktabs}
\usepackage{dblfloatfix}
\usepackage{cite}
\usepackage{amsmath}
\usepackage{amssymb}
\usepackage{xcolor}
\usepackage[colorlinks=true, allcolors=black]{hyperref}
\newcommand{\tX}{\Tilde{X}}
\newcommand{\tx}{\Tilde{x}}

\begin{document}
\title{Application of the representative measure approach to assess the reliability of decision trees in dealing with unseen vehicle collision data}
\titlerunning{Representative measure approach to assess the reliability of decision trees}
\author{Javier Perera-Lago\inst{1}\orcidID{0009-0009-4536-4020} \and \\
Víctor Toscano-Durán\inst{1}\orcidID{0009-0006-1316-9026} \and \\
Eduardo Paluzo-Hidalgo\inst{1,2}\orcidID{0000-0002-4280-5945} \and \\
Sara Narteni\inst{3}\orcidID{0000-0002-0579-647X}\and \\
Matteo Rucco\inst{4}\orcidID{0000-0003-2561-3340}}
\authorrunning{J. Perera et al.}
\institute{Department of Applied Mathematics I, University of Sevilla, Sevilla, Spain \email{\{jperera,vtoscano\}@us.es}\and
Department of Quantitative Methods, Universidad Loyola Andalucía, Campus Sevilla, Dos Hermanas, Seville, Spain\\ \email{epaluzo@\{uloyola,us\}.es}\and
Cnr-Istituto di Elettronica, Ingegneria
dell’Informazione e delle Telecomunicazioni (CNR-IEIIT), Genoa, Italy\\
\email{sara.narteni@ieiit.cnr.it}
\and
Data Science Department, Biocentis, Milan, Italy\\
\email{matteo.rucco@biocentis.com}}
\maketitle
\begin{abstract}
Machine learning algorithms are fundamental components of novel data-informed Artificial Intelligence architecture. In this domain, the imperative role of representative datasets is a cornerstone in shaping the trajectory of artificial intelligence (AI) development. Representative datasets are needed to train machine learning components properly. Proper training has multiple impacts: it reduces the final model's complexity, power, and uncertainties. In this paper, we investigate the reliability of the $\varepsilon$-representativeness method to assess the dataset similarity from a theoretical perspective for decision trees. We decided to focus on the family of decision trees because it includes a wide variety of models known to be explainable. Thus, in this paper, we provide a result guaranteeing that if two datasets are related by $\varepsilon$-representativeness, i.e., both of them have points closer than $\varepsilon$, then the predictions by the classic decision tree are similar. Experimentally, we have also tested that $\varepsilon$-representativeness presents a significant correlation with the ordering of the feature importance. Moreover, we extend the results experimentally in the context of unseen vehicle collision data for XGboost, a machine-learning component widely adopted for dealing with tabular data.
\end{abstract}
\section{Introduction}

In the contemporary landscape of technological evolution, the imperative role of representative datasets stands as a cornerstone in shaping the trajectory of artificial intelligence (AI) development. As AI algorithms increasingly influence decision-making processes, the need for robust, unbiased, and comprehensive datasets becomes ever more critical \cite{liang2022advances}.
The frameworks that provide the guidelines and the technical requirements for engineering trustworthy data-driven AI systems provide at least two possible interpretations of 
representativeness \cite{floridi2018soft, diakopoulos2016accountability, arrieta2020explainable, jobin2019global, liu2019towards, brundage2021toward, wachter2017counterfactual}: 1) attribute coverage and completeness: how much the samples in a dataset describe the phenomena under observations. The question mark can be interpreted both in terms of the number of features that have been recorded in the dataset and the number and sparsity of samples concerning the dynamics of the system. 2) similarity among datasets: given two or more datasets, how much are they different, and what is their impact on a data-driven model derived from the datasets. Attribute coverage and completeness can be measured with different approaches. Methods such as Population Parity and Disparate Impact quantify and identify disparities in the distribution of different demographic groups within a dataset \cite{pedreschi2008discrimination, zafar2017fairness}. Statistical techniques like Feature Divergence and Kernel Density Estimation can be used to assess the similarity between feature distributions in the dataset and the real-world population \cite{sugiyama2012density, scott2015multivariate}. Fairness Indicators and Counterfactual Fairness detect and mitigate biases in AI models trained on biased datasets \cite{hardt2016equality, zemel2013learning}. Outlier Analysis and Clustering Techniques can be used to identify the points that might compromise dataset representativeness \cite{chandola2009anomaly,kriegel2009outlier}.  Temporal Drift Detection and Cohort Analysis are explored, shedding light on techniques that ensure the dataset's representativeness remains intact as conditions evolve over time  \cite{widmer1996learning, rahman2019detecting}. Transfer Learning and Adversarial Training are discussed in adapting models to diverse domains, mitigating biases associated with domain-specific features \cite{pan2010survey, ganin2016domain}.  Crowdsourcing and Expert Evaluation are presented as methods to incorporate diverse perspectives, providing insights into potential biases and limitations in the dataset from end-users and domain experts \cite{kittur2008harnessing, kittur2013crowdsourcing}.  Intersectional Approaches and Subgroup Analysis are considered, emphasizing the importance of examining the interaction between multiple demographic attributes for a more comprehensive assessment of representativeness \cite{crenshaw1989demarginalizing}. 

On the other hand, assessing the similarity of datasets is crucial for ensuring that machine learning models trained on these datasets generalize well to real-world scenarios. Various methods have been developed to measure the likeness between datasets, to identify potential discrepancies, and to ensure that the data used for model training accurately represents the target domain. Metrics such as Wasserstein distance, Jensen-Shannon divergence, and Bhattacharyya distance offer quantitative measures of dissimilarity between probability distributions, providing insights into the overall similarity of datasets \cite{rubner2000earth, bhattacharyya1943measure}. Additionally, domain adaptation techniques, such as Maximum Mean Discrepancy (MMD) and Kernelized Discrepancy (KMM), focus on aligning feature distributions between source and target domains, ensuring a seamless transition from one dataset to another \cite{gretton2007kernel,huang2007correcting}. Understanding and mitigating dissimilarities between datasets are paramount to building robust and generalizable machine learning models, as models trained on dissimilar datasets may exhibit poor performance when deployed in real-world applications. Therefore, a comprehensive assessment of dataset similarity is indispensable for fostering reliable and effective AI systems. In this paper, 
we use the measure proposed in \cite{gonzalez2022topology} to quantify the
similarity between datasets and how their difference has an impact on a machine learning component, decision trees and, in particular, for the eXtreme Gradient Boosting (aka XGBoost). 
XGBoost stands out as a powerful and versatile machine learning algorithm with significant implications for trustworthy AI. Its importance lies in its ability to enhance model performance across various tasks, such as classification, regression, and ranking. XGBoost excels in handling complex datasets, mitigating overfitting, and providing robust predictions. The algorithm's interpretability features, such as the ability to generate feature importance scores and decision trees, contribute to the transparency of AI models—a crucial aspect for ensuring trustworthiness. Moreover, XGBoost's regularized learning objectives and advanced optimization techniques foster model generalization and prevent overfitting, reinforcing the reliability of AI systems. The algorithm's widespread adoption in both research and industry attests to its effectiveness and underscores its role in building trustworthy AI models that prioritize accuracy, interpretability, and generalization \cite{chen2016xgboost}.

This paper is organized as follows. In Section~\ref{sec:background}, the main concepts about decision trees are provided. Then, in Section~\ref{sec:theoretical_results}, we discuss the relationship between the $\varepsilon$-representativeness measure and the predictions of decision trees proving theoretical results. In Section~\ref{sec:experiments}, we explore experimentally the correlation between $\varepsilon$-representativeness and the similarity between the explanations provided by decision trees and XGBoost by the ordering of the feature importance. Finally, conclusions and future work are discussed in Section~\ref{sec:conclusions}.

\section{Classification with decision trees and XGBoost}\label{sec:background}

In this section, we introduce all the needed concepts about classification and the model family of decision trees. For a further understanding of machine learning, we refer to \cite{MohriRostamizadehTalwalkar18}, and for geometric results for point clouds we refer to \cite{Boissonnat_Chazal_Yvinec_2018}.

Let $(X,\lambda_X)$ be a dataset for classification with $X\subseteq \mathbb{R}^d$ the set of data with size $N$ and $\lambda_X : X \rightarrow \{1,\dots, c\}$ the class labelling. For each $x\in X$, the $j$-th coordinate coordinate of $x$ will be denoted by $x_j \in \mathbb{R}$ 
and will be called the $j$-th feature of $x$. 
The objective of classification is to find a function $f:\mathbb{R}^n\rightarrow \{1,\dots, c\}$ that approximates $\lambda_X$. In the literature, there exist different families of functions $f$, such as artificial neural networks and support vector machines, which are called models \cite{MohriRostamizadehTalwalkar18}. We will focus on the family of decision trees $\mathcal{T}$ which are supervised learning models with desirable properties such as that they are interpretable by definition. The most basic model of the family is decision trees (DT) (see \cite[Section 9.3.3]{MohriRostamizadehTalwalkar18}). 

A binary DT is a rooted tree representing a partition of the feature space. It contains a set of ordered nodes $\{n_1,\cdots,n_r\}$ where $n_1$ will be called the root node. A node is said to be internal if it is connected to two children nodes, opening new branches in the tree. Conversely, terminal nodes, also known as leaves, do not have children and represent the endpoints of a branch in the tree. Let $I \subset \{1, \cdots, r\}$ be the subset of indices corresponding to the internal nodes, and $L$ be the subset of indices corresponding to the terminal nodes. The root node $n_1$ is considered an internal node, so $1 \in I$.

Given a data $x\in X$, $x$ traverses the internal nodes of the binary DT from the root $n_1$ to one of the terminal nodes.
Each internal node $n_i$ is associated with one of the features $j_i\in \{1,\cdots,n\}$, and with one condition in terms of an inequality bounded by a threshold value $t_i\in \mathbb{R}$ that will send the data to one of the two children nodes based on the inequality. This inequality is called a decision rule.
When $x$ reaches an internal node $n_i$, it is sent to its left child if $t_i-x_{j_i} > 0$ and to its right child otherwise. The margin $\mu_i > 0$ of an internal node $n_i$ is the minimum value $|t_i-x_{j_i}|$ for all the examples $x \in X$ reaching $n_i$. On the other hand, each terminal node $n_{\ell}$ is associated with an integer $k_\ell \in \{1,\cdots,c\}$ representing a class label. When $x$ reaches a terminal node $n_{\ell}$, it finishes its path through the binary DT and is predicted to be in the class $k_{\ell}$.

There exist different training algorithms for binary DTs. In our case, let us consider first a binary DT with just one node that splits the dataset into two subsets based on one of the features. The feature and the splitting condition are chosen based on minimizing the impurity of the nodes. The impurity of a node is a measure of the homogeneity of the class labels of the examples that reach it. A node has maximum impurity when all the class labels are evenly distributed, and has minimum impurity (it is called a pure node) when all the labels are the same. Then, the process is iterated recursively until a desired depth is reached or the impurity can not be improved. There exist different impurity measures. The most common ones are the entropy and the Gini index \cite[Section 9.3.3]{MohriRostamizadehTalwalkar18}. 

Let us denote by $N_i$ the number of examples from $X$ reaching the node $n_i$. The number of class $k$ examples reaching $n_i$ will be denoted by $N_{i,k}$. Then, let us use the following notation: $p_i = N_i/N$, $p_{i,k}=N_{i,k}/N_i$. The entropy of a node $n_i$ is $E(n_i) =-\sum_{k=1}^c p_{i,k}\log{p_{i,k}}$, and the Gini index of $n_i$ is $G(n_i) =\sum_{k=1}^c p_{i,k}(1-p_{i,k})$. Assume that we fix an impurity measure and we denote it as $I$. The information gain for an internal node $n_i$ whose two children nodes are $n_{i_1}$ and $n_{i_2}$ is:

$$IG(n_i) = I(n_i) - \frac{N_{i_1}}{N_{i}}I(n_{i_1}) - \frac{N_{i_2}}{N_{i}}I(n_{i_2})$$

Feature importance ($FI$) quantifies the impact of a particular feature $j\in\{1,\cdots,d\}$ in decreasing
the impurity of the decision tree. It is calculated as:

$$FI(j) = \sum_{\substack{i \in I \\ j_i = j}} N_i \cdot IG(n_i)$$

The goodness of the classification given by $T \in \mathcal{T}$ for the dataset $(X,\lambda_X)$ is measured by the accuracy, with formula
$\operatorname{Acc}(T,(X,\lambda_X)) = \sum_{\ell\in L} p_{\ell} \cdot p_{\ell,k_{\ell}}$.

A more complex algorithm belonging to $\mathcal{T}$ is called XGBoost~\cite{xgboost}. A thorough description of XGBoost is out of the scope of this paper but, roughly speaking, it is an ensemble of decision trees, i.e., it combines the predictions of several binary DTs that are sequentially built, each correcting errors of the previous one. Then, during the training a gradient descent optimization algorithm is used.

\section{Representativeness and decision trees}\label{sec:theoretical_results}

In this section, we introduce a metric to compare different training datasets and we show a theoretical result proving the relation between this metric and the final performance of a binary DT. 

Given a dataset $(X,\lambda_X)$, the ability of a binary DT to predict new unseen data will depend on the completeness and quality of the information learned during training. It becomes then a vital task to find measures to compare datasets expected to induce similar predictions. In~\cite{gonzalez2022topology}, a measure based on computational topology was proposed. This measure is called the $\varepsilon$-representativeness of a dataset. Given another dataset $(\tX,\lambda_{\tX})$ with the cardinality of $\tX$ smaller than the one of $X$, we say that $\tx \in \tX$ is an $\varepsilon$-representative of $x \in X$ if $||\tx-x||_{\infty}\le \varepsilon$ and $\lambda_X(x)=\lambda_{\tX}(\tx)$, and we say that $(\tX,\lambda_{\tX})$ is an $\varepsilon$-representative dataset of $(X,\lambda_X)$ if for all $x\in X$ there exists $\tx\in \tX$ that is an $\varepsilon$-representative of $x$. In general, we will consider $\tX$ to be a subset of $X$ but it is not a necessary condition. All the notations that we defined in the previous section for $(X,\lambda_X)$ will have the symbol \ $\Tilde{}$ \ when we refer to $(\tX,\lambda_{\tX})$ (for example, we will say that the size of $\tX$ is $\Tilde{N}$).
A dataset $(\tX,\lambda_{\tX})$ that is representative of $(X,\lambda_X)$ is said to be $\gamma$-balanced if each $\tx \in \tX$ is representative of exactly $\gamma$ data examples of $X$ and each $x \in X$ is represented by a single example $\tx \in \tX$.

Knowing the concept of representativeness, we present the following result:

\begin{theorem}\label{th:theorem1}
Let $T \in \mathcal{T}$ be a binary DT, $(X,\lambda_X)$ a dataset and $(\tX,\lambda_{\tX})$ a $\gamma$-balanced $\varepsilon$-representative dataset of $(X,\lambda_X)$. If $\varepsilon < M = \min_{i \in I} \mu_i$, then $$\operatorname{Acc}(T,(X,\lambda_X)) = \operatorname{Acc}(T,(\tX,\lambda_{\tX}))$$
\end{theorem}
\proof
Let $x = (x_1,\cdots,x_n)^T \in X$ and $\tx = (\tx_1,\cdots,\tx_n)^T \in \tX$ be an $\varepsilon$-representative of $x$. That means that $|\tx_j-x_j| \leq \varepsilon \ \forall j\in \{1,\cdots,d\}$. Assume that $\tx$ accesses the tree through the root node $n_1$ and is sent to its left child. That means that $0 < t_1 - \tx_{j_1}$. By the definition of margins, we know that $ \mu_1 \leq t_1 - \tx_{j_1}$. Since $\tx$ is $\varepsilon$-representative of $x$, we have that $x_{j_1} \leq \tx_{j_1}+\varepsilon$. Adding these two last inequalities,
we have that $\mu_1 - \varepsilon \leq t_1 - x_{j_1}$. Since $\varepsilon < M \leq \mu_1$, then $0 < t_1 - x_{j_1}$, meaning that $x$ is also sent to the left child. Analogously, if we assume that $\tx$ is sent to the right child of $n_1$, we can show that $x$ is also sent to the right child. 
If we apply this same reasoning to all the internal nodes that $\tx$ passes through, we see that $x$ follows the same path through the tree and, consequently, reaches the same terminal node $n_\ell$ and is classified with the same label $k_\ell$.

By the definition of $\gamma$-balance, for each $\tx \in \tX$ of class $k$ there are exactly $\gamma$ examples from $X$ of class $k$ $\varepsilon$-represented by $\tx$, and all of them reach the same terminal node of $T$. It also follows from the definition of $\gamma$-balance that $N = \gamma \cdot \Tilde{N}$, $N_j=\gamma \cdot \Tilde{N}_j$ and $N_{j,k}=\gamma \cdot \Tilde{N}_{j,k}$. Consequently, $p_j = N_j/N = (\gamma \cdot \Tilde{N}_j)/(\gamma \cdot \Tilde{N}) = \Tilde{N}_j/\Tilde{N} = \Tilde{p}_j$ and $p_{j,k} = N_{j,k}/N_j = (\gamma \cdot \Tilde{N}_{j,k})/(\gamma \cdot \Tilde{N}_j) = \Tilde{N}_{j,k}/\Tilde{N}_j = \Tilde{p}_{j,k}$. Consequently: $$
\operatorname{Acc}(T,(X,\lambda_X)) = \sum_{j\in L} p_j \cdot p_{j,k_{j}} = \sum_{j\in L} \Tilde{p}_j \cdot \Tilde{p}_{j,k_{j}} = \operatorname{Acc}(T,(\tX,\lambda_{\tX}))$$ \qed

It is important to note that this result pertains to the accuracy of a fixed binary DT when evaluated on both the full dataset and a representative subset. We cannot guarantee that training will yield a similar DT when using either the full dataset or its representative subset. However, if the accuracy is high for the DT trained on the representative dataset, Theorem~\ref{th:theorem1} ensures that its accuracy will also be high on the full dataset.

\section{Experiments}\label{sec:experiments}

The experiments developed are organized as follows. Firstly, a $2D$ synthetic dataset was used as an illustration of the proposed methodology. Then, the experimentation was extended to the vehicle platooning (also called collision) dataset~\cite{collision}. Given a dataset, subsets of the set with different values of $\varepsilon$-representativeness will be considered and the ordering of the features based on the importance compared. The results suggest that the lower values of $\varepsilon$ will produce similar explanations of the input data and similar decision boundaries. The code for the experiments is available in a GitHub repository \footnote{\url{https://github.com/Cimagroup/Application_Representative_Measure_Reliability_DT}}. 

\subsection{Synthetic $2D$ dataset}\label{sec:synthetic}

\begin{figure}[h!]
    \centering
    \includegraphics[width=0.32\textwidth,height=0.4\textwidth]{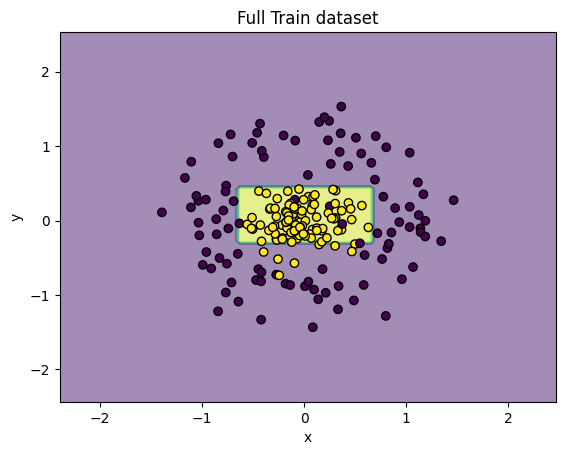}
    \includegraphics[width=0.32\textwidth,height=0.4\textwidth]{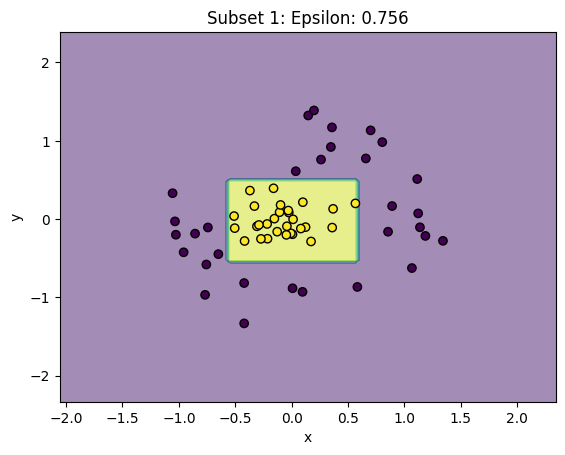}
    \includegraphics[width=0.32\textwidth,height=0.4\textwidth]{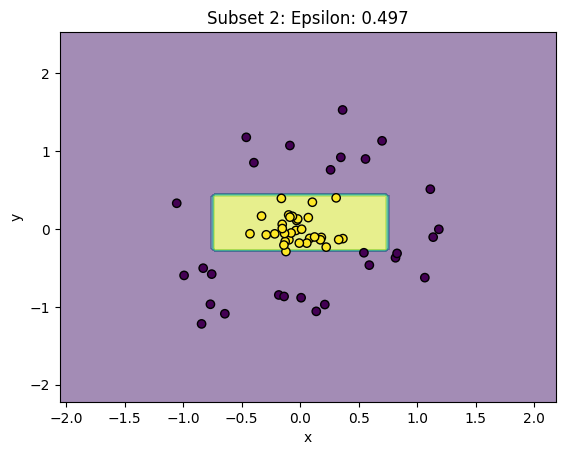}
    \caption[Full synthetic dataset generated using Scikit-learn and the two random subsets.]{\label{representationDataset} Full synthetic dataset generated using Scikit-learn and the two random subsets. From left to right: (1) the training set; (2) a subset composed of a $40\%$ of the training set and $\varepsilon=0.756$; (3) a subset composed of $40\%$ of the training set and $\varepsilon=0.497$. We can also see the decision boundaries of the binary DTs trained using each set of data.}
\end{figure}

In this experiment, we used a $2D$ dataset generated using the Python Scikit-learn package\footnote{\url{https://scikit-learn.org/stable/index.html}} as an example. The dataset comprises $200$ data distributed in two noisy concentric circles representing distinct classes as shown in Figure~\ref{representationDataset}.

The methodology followed in the experiments is summarized in the following steps. The dataset was split into a training set composed of the $75\%$ of the data and a test set with the remaining $25\%$. Then, two random subsets containing the $40\%$ of the training set were considered, from now on Subset 1 and Subset 2, and their $\varepsilon$-representativeness was computed, obtaining $\varepsilon=0.756$ for the Subset 1 and $\varepsilon=0.497$ for the Subset 2. 

\begin{figure}[h!]
    \centering     \includegraphics[width=0.5\textwidth]{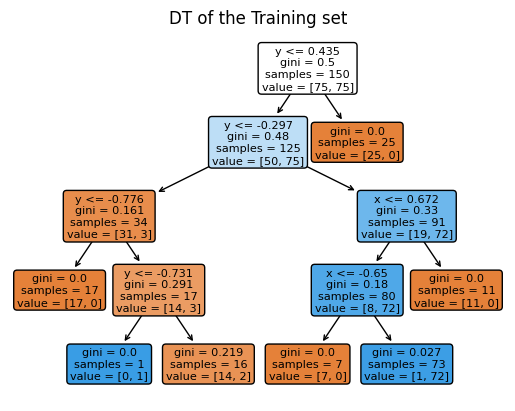}
    \includegraphics[width=0.5\textwidth]{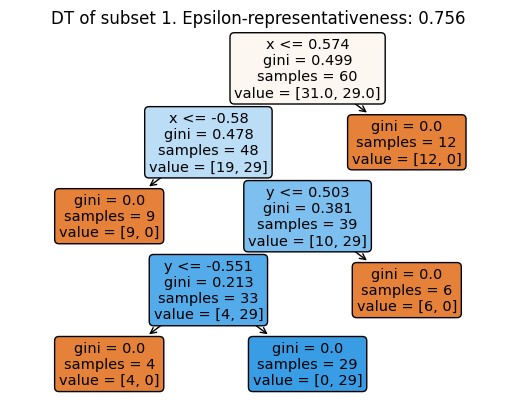}
    \includegraphics[width=0.5\textwidth]{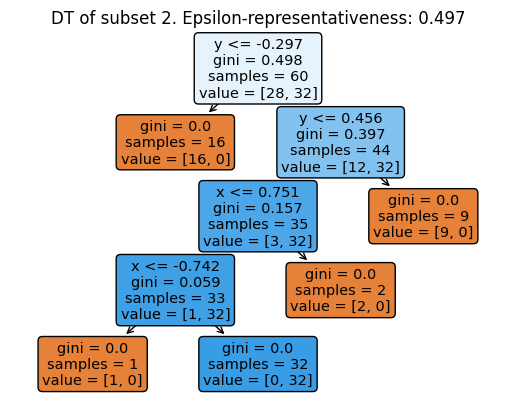}\caption{\label{decisionTreesSynthetic} Binary DTs for the sets in Figure~\ref{representationDataset}. From top to bottom: (1) the training set; (2) a subset composed of a $40\%$ of the training set and $\varepsilon=0.756$; (3) a subset composed of $40\%$ of the training set and $\varepsilon=0.497$.}
\end{figure}

A binary DT was trained with the training set and the subsets using the Gini index and a maximum depth of $4$ obtaining the binary DTs of Figure~\ref{decisionTreesSynthetic}. As we can see, the resultant decision rules after training the DT with the training set are similar with a lower value of $\varepsilon$. The accuracy values for the binary DTs on the test set were: $0.84$ for the train dataset, $0.94$ for Subset 1, and $0.82$ for Subset 2. To determine the similarities between the binary DTs, we considered the ordering of the feature importance. The first feature ranked the most important for both the training set and Subset 2 while the ordering was the opposite for Subset 1 (See Table~\ref{tab:feature_importance_synthetic}).  

\begin{table}[h]
\centering
\caption{Feature importance percentage for the training set, Subset 1, and Subset 2. We can see that the most important feature for the training set and Subset 2 is the same.}
\label{tab:feature_importance_synthetic}
\begin{tabular}{@{}lcc@{}}
\toprule
\textbf{Data} & \textbf{$x_1$} & \textbf{$x_2$} \\ \midrule
Training set  & 40.4            & \textbf{59.6}  \\
Subset 1      & \textbf{50.37}  & 49.62          \\
Subset 2      & 18.4            & \textbf{81.6}  \\ \bottomrule
\end{tabular}
\end{table}

\subsection{The collision dataset}

In this experiment, we used the binary classification collision Dataset \cite{collision} which consists of predicting whether a platoon of vehicles will collide based on features such as the number of cars or their speed. It is composed of 107,210 data with 23 numerical features. Each class is composed of 69,348 examples and 37,862 examples, respectively.

We followed a similar methodology to the one proposed for the experiment in Subsection~\ref{sec:synthetic}. The dataset was split into a training set composed of the $75\%$ of the data and a test set with the remaining $25\%$. Then, two random subsets containing the $10\%$ of the training data were considered, denominated as Subset 1 and Subset 2, and their $\varepsilon$-representativeness was computed, obtaining $\varepsilon=0.539$ for Subset 1 and $\varepsilon=0.655$ for Subset 2. A binary DT was trained with the training set and the subsets using the Gini index and a maximum depth of 10. The following accuracy values were obtained on the test set: $0.874$ for the training dataset, $0.841$ for Subset 1, and $0.84$ for Subset 2. 

To determine the similarities between the binary DTs, we considered the ordering of the feature importance. For comparing the ordering of the feature importance, we used a metric developed in \cite[Section 4.2]{Barrera-VicentP23}. Let $x$ and $y$ be two ordered sets whose elements are the features of the dataset ordered by their importance. Then, for each feature, we compute the absolute value of the difference between the position of the feature in $x$ and $y$. Finally, the mean of these differences is computed. For Subset 1, we obtained a value of $1.74$ for this metric, and for Subset 2 the value was $1.83$, where we can highlight the higher similarity of variable importance for the decision tree generated by the subset with lower epsilon, that is the Subset 1. The ordering of the feature importance for the three binary DTs is displayed in Table~\ref{tab:importance_order}. Finally, the experiment was repeated for 100 subsets and the Spearman's correlation ($Sp$) between the $\varepsilon$-representativeness and the metric of the ordering of the feature importance was computed obtaining significant correlation ($Sp=0.51$, $p$-value$=5.2\times 10^{-8}$).

\begin{table}[h!]
\centering
\caption{Ordering of the feature importance for binary DTs and XGBoost trained with the training set and the random subsets. The number indicates the position in the importance order. For example, the most important feature for binary DT trained on the training set is Int\_dv7.}
\begin{tabular}{cccc|ccc}
\toprule
 & \multicolumn{3}{c}{\textbf{DT}} &  \multicolumn{3}{c} {\textbf{XGBoost}} \\
\midrule
\textbf{} & \textbf{Training set} & \textbf{Subset 1} & \textbf{Subset 2} & \textbf{Training set} & \textbf{Subset 1} & \textbf{Subset 2} \\
\midrule
\textbf{F0} & 2 & 4 & 2 & 2 & 3 & 2 \\
\textbf{d\_ms} & 10 & 10 & 10 & 10 & 10 & 10 \\
\textbf{d0} & 6 & 8 & 9 & 5 & 7 & 9 \\
\textbf{v0} & 23 & 23 & 20 & 23 & 23 & 23 \\
\textbf{prob} & 5 & 5 & 8 & 6 & 5 & 7 \\
\textbf{Int\_dv1} & 15 & 14 & 12 & 13 & 13 & 12 \\
\textbf{Int\_dv2} & 12 & 15 & 19 & 16 & 15 & 21 \\
\textbf{Int\_dv3} & 14 & 17 & 14 & 14 & 14 & 13 \\
\textbf{Int\_dv4} & 7 & 9 & 7 & 9 & 9 & 8 \\
\textbf{Int\_dv5} & 20 & 12 & 21 & 18 & 16 & 19 \\
\textbf{Int\_dv6} & 16 & 16 & 23 & 15 & 17 & 20 \\
\textbf{Int\_dv7} & 1 & 1 & 1 & 1 & 1 & 1 \\
\textbf{Int\_dd2} & 18 & 22 & 16 & 19 & 19 & 14 \\
\textbf{Int\_dd3} & 19 & 18 & 15 & 20 & 20 & 16 \\
\textbf{Int\_dd4} & 3 & 3 & 3 & 4 & 4 & 4 \\
\textbf{Int\_dd5} & 13 & 13 & 13 & 12 & 12 & 15 \\
\textbf{Int\_dd6} & 17 & 21 & 17 & 17 & 18 & 18 \\
\textbf{Int\_dd7} & 11 & 11 & 11 & 11 & 11 & 11 \\
\textbf{duration} & 22 & 20 & 22 & 21 & 22 & 22 \\
\textbf{freq} & 4 & 2 & 4 & 3 & 2 & 3 \\
\textbf{ampl} & 8 & 7 & 6 & 7 & 8 & 6 \\
\textbf{Fresponse} & 21 & 19 & 18 & 22 & 21 & 17 \\
\textbf{KInt} & 9 & 6 & 5 & 8 & 6 & 5 \\
\bottomrule
\end{tabular}
\label{tab:importance_order}
\end{table}

Additionally, we trained Gradient Boosting Classifier(XGBoost) with the training set and the subsets using the Friedman Mean Squared Error as a criterion and a maximum depth of 10. The number of boosting stages to perform was set to 25. The following accuracy values were obtained: $0.912$ for the training dataset, $0.882$ for Subset 1, and $0.876$ for Subset 2. In this case, Subset 1 has a similarity of $0.696$ for the feature importance, and Subset 2 has a similarity value of $1.823$, reaching better similarity with lower $\varepsilon$. The ordering of the feature importance for the three XGBoost trained models is displayed in Table~\ref{tab:importance_order}. Finally, the experiment was repeated for 100 subsets and the Spearman's correlation ($Sp$) between the $\varepsilon$-representativeness and the metric of the ordering of the feature importance was computed obtaining significant correlation ($Sp=0.673$, $p$-value$=1.79\times 10^{-14}$).

\section{Conclusions and future work}\label{sec:conclusions}

The results of this paper are two-fold. Firstly, we proved that similar accuracy is obtained when certain conditions about representativeness are satisfied when using binary DTs. Secondly, by experimentation, we have compared the feature importance ordering for different subsets of the training data with different values of $\varepsilon$-representativeness reaching a significant correlation between them. According to our results, representative sets produce similar explanations of the dataset. Finally, we extend our experiments to a more complex decision tree algorithm, XGBoost. In the future, we would like to provide theoretical guarantees regarding the ordering of the importance of the features and distance-based comparison between the decision rules of binary DTs.

\section*{Acknowledgements}
We want to thank Maurizio Mongelli and Miguel A. Gutierrez-Naranjo for the insightful discussions and ideas. The work was supported in part by the European Union HORIZON-CL4-2021-HUMAN-01-01 under grant agreement 101070028 (REXASI-PRO) and by  
TED2021-129438B-I00 / AEI/10.13039/501100011033 / Unión Europea NextGenerationEU/PRTR.

\bibliographystyle{splncs04}
\bibliography{biblio}

\begin{thebibliography}{10}
\providecommand{\url}[1]{\texttt{#1}}
\providecommand{\urlprefix}{URL }
\providecommand{\doi}[1]{https://doi.org/#1}

\bibitem{arrieta2020explainable}
Arrieta, A.B., Díaz-Rodríguez, N., Del~Ser, J., Bennetot, A., Tabik, S., Barbado, A., Garcia, S., Gil-Lopez, S., Molina, D., Benjamins, R., et~al.: Explainable artificial intelligence (xai): Concepts, taxonomies, opportunities and challenges toward responsible ai. In: 2019 IEEE international conference on systems, man and cybernetics (SMC). pp. 4184--4191. IEEE (2020)

\bibitem{Barrera-VicentP23}
Barrera-Vicent, A., Paluzo-Hidalgo, E., Gutiérrez-Naranjo, M.A.: The metric-aware kernel-width choice for lime. In: Longo, L. (ed.) Joint Proceedings of the xAI-2023 Late-breaking Work, Demos and Doctoral Consortium co-located with the 1st World Conference on eXplainable Artificial Intelligence (xAI-2023), Lisbon, Portugal, July 26-28, 2023. CEUR Workshop Proceedings, vol.~3554, pp. 117--122. CEUR-WS.org (2023), \url{https://ceur-ws.org/Vol-3554/paper21.pdf}

\bibitem{bhattacharyya1943measure}
Bhattacharyya, A.: A measure of asymptotic efficiency for tests of a hypothesis based on the sum of observations. Biometrika  \textbf{34}(3/4),  291--302 (1943)

\bibitem{Boissonnat_Chazal_Yvinec_2018}
Boissonnat, J.D., Chazal, F., Yvinec, M.: Geometric and Topological Inference. Cambridge Texts in Applied Mathematics, Cambridge University Press (2018)

\bibitem{brundage2021toward}
Brundage, e.a.: Toward trustworthy ai development: Mechanisms for supporting verifiable claims. arXiv preprint arXiv:2110.05282  (2021)

\bibitem{chandola2009anomaly}
Chandola, V., Banerjee, A., Kumar, V.: Anomaly detection: A survey. ACM Computing Surveys (CSUR)  \textbf{41}(3),  1--58 (2009)

\bibitem{chen2016xgboost}
Chen, T., Guestrin, C.: Xgboost: A scalable and accurate implementation of gradient boosting. Proceedings of the 22nd ACM SIGKDD International Conference on Knowledge Discovery and Data Mining pp. 785--794 (2016)

\bibitem{xgboost}
Chen, T., Guestrin, C.: Xgboost: A scalable tree boosting system. In: Proceedings of the 22nd ACM SIGKDD International Conference on Knowledge Discovery and Data Mining. p. 785–794. KDD '16, Association for Computing Machinery, New York, NY, USA (2016). \doi{10.1145/2939672.2939785}, \url{https://doi.org/10.1145/2939672.2939785}

\bibitem{crenshaw1989demarginalizing}
Crenshaw, K.: Demarginalizing the intersection of race and sex: A black feminist critique of antidiscrimination doctrine, feminist theory and antiracist politics. University of Chicago Legal Forum pp. 139--167 (1989)

\bibitem{diakopoulos2016accountability}
Diakopoulos, N.: Accountability in algorithmic decision making: A visual analysis framework. Data Society Research Institute  (2016)

\bibitem{floridi2018soft}
Floridi, L., Taddeo, M.: Soft ethics, the governance of the digital and the general data protection regulation. Philosophical Transactions of the Royal Society A: Mathematical, Physical and Engineering Sciences  \textbf{376}(2133),  20180081 (2018)

\bibitem{ganin2016domain}
Ganin, Y., Ustinova, E., Ajakan, H., Germain, P., Larochelle, H., Laviolette, F., Lempitsky, V.: Domain-adversarial training of neural networks. Journal of Machine Learning Research  \textbf{17}(1),  2096--2030 (2016)

\bibitem{gonzalez2022topology}
Gonzalez-Diaz, R., Guti{\'e}rrez-Naranjo, M.A., Paluzo-Hidalgo, E.: Topology-based representative datasets to reduce neural network training resources. Neural Computing and Applications  (2022)

\bibitem{gretton2007kernel}
Gretton, A., Borgwardt, K.M., Rasch, M.J., Sch{\"o}lkopf, B., Smola, A.: A kernel method for the two-sample-problem. In: Advances in neural information processing systems. pp. 513--520 (2007)

\bibitem{hardt2016equality}
Hardt, M., Price, E., Srebro, N.: Equality of opportunity in supervised learning. In: Advances in neural information processing systems. pp. 3315--3323 (2016)

\bibitem{huang2007correcting}
Huang, J., Smola, A.J., Gretton, A., Borgwardt, K.M., Sch{\"o}lkopf, B.: Correcting sample selection bias by unlabeled data. Advances in neural information processing systems  \textbf{19}, ~601 (2007)

\bibitem{jobin2019global}
Jobin, A., Ienca, M., Vayena, E.: Global data justice. Ethics and Information Technology  \textbf{21}(2),  87--96 (2019)

\bibitem{kittur2013crowdsourcing}
Kittur, A., Chi, E.H., Suh, B.: Crowdsourcing user studies with mechanical turk. In: Proceedings of the SIGCHI Conference on Human Factors in Computing Systems. pp. 453--456 (2013)

\bibitem{kittur2008harnessing}
Kittur, A., Kraut, R.E.: Harnessing the wisdom of crowds in wikipedia: Quality through coordination. In: Proceedings of the 2008 ACM conference on Computer supported cooperative work. pp. 37--46 (2008)

\bibitem{kriegel2009outlier}
Kriegel, H.P., Kr{\"o}ger, P., Schubert, E., Zimek, A.: Outlier detection in axis-parallel subspaces of high-dimensional data. In: Proceedings of the 18th ACM conference on Information and knowledge management. pp. 555--564 (2009)

\bibitem{liang2022advances}
Liang, W., Tadesse, G.A., Ho, D., Fei-Fei, L., Zaharia, M., Zhang, C., Zou, J.: Advances, challenges and opportunities in creating data for trustworthy ai. Nature Machine Intelligence  \textbf{4}(8),  669--677 (2022)

\bibitem{liu2019towards}
Liu, Y., Zhao, Z., Guan, Y., Song, S., Cao, L., Chen, H.: Towards trustworthy ai: A cross-disciplinary survey. arXiv preprint arXiv:1907.02527  (2019)

\bibitem{MohriRostamizadehTalwalkar18}
Mohri, M., Rostamizadeh, A., Talwalkar, A.: Foundations of machine learning. MIT press (2018)

\bibitem{collision}
Mongelli, M., Ferrari, E., Muselli, M., Fermi, A.: Performance validation of vehicle platooning through intelligible analytics. IET Cyber-Physical Systems: Theory \& Applications  \textbf{4}(2),  120--127 (2019). \doi{https://doi.org/10.1049/iet-cps.2018.5055}, \url{https://ietresearch.onlinelibrary.wiley.com/doi/abs/10.1049/iet-cps.2018.5055}

\bibitem{pan2010survey}
Pan, S.J., Yang, Q.: A survey on transfer learning. IEEE Transactions on knowledge and data engineering  \textbf{22}(10),  1345--1359 (2010)

\bibitem{pedreschi2008discrimination}
Pedreschi, D., Ruggieri, S., Turini, F.: Discrimination-aware data mining. In: Proceedings of the 14th ACM SIGKDD international conference on Knowledge discovery and data mining. pp. 560--568 (2008)

\bibitem{rahman2019detecting}
Rahman, S.M.M.M., Davis, D.: Detecting and mitigating concept drift in an adaptive learning system for personalized news recommendation. Knowledge-Based Systems  \textbf{166},  132--145 (2019)

\bibitem{rubner2000earth}
Rubner, Y., Tomasi, C., Guibas, L.J.: The earth mover's distance as a metric for image retrieval. International Journal of Computer Vision  \textbf{40}(2),  99--121 (2000)

\bibitem{scott2015multivariate}
Scott, D.W.: Multivariate density estimation: theory, practice, and visualization. John Wiley \& Sons (2015)

\bibitem{sugiyama2012density}
Sugiyama, M., Suzuki, T., Kanamori, T.: Density-ratio matching under the bregman divergence: a unified framework of density-ratio estimation. Annals of the Institute of Statistical Mathematics  \textbf{64}(5),  1009--1044 (2012)

\bibitem{wachter2017counterfactual}
Wachter, S., Mittelstadt, B., Russell, C.: Counterfactual explanations without opening the black box: Automated decisions and the gdpr. In: 2017 IEEE international conference on data science and advanced analytics (DSAA). pp. 520--529. IEEE (2017)

\bibitem{widmer1996learning}
Widmer, G., Kubat, M.: Learning in the presence of concept drift and hidden contexts. Machine learning  \textbf{23}(1),  69--101 (1996)

\bibitem{zafar2017fairness}
Zafar, M.B., Valera, I., Rodriguez, M.G., Gummadi, K.P.: Fairness beyond disparate treatment \& disparate impact: Learning classification without disparate mistreatment. In: Proceedings of the 26th International Conference on World Wide Web. pp. 1171--1180 (2017)

\bibitem{zemel2013learning}
Zemel, R., Wu, Y., Swersky, K., Pitassi, T., Dwork, C.: Learning fair representations. In: Proceedings of the 30th International Conference on Machine Learning (Vol. 28, No. 3) (2013)

\end{thebibliography}

\end{document}